\documentclass[sigconf]{acmart}
\usepackage{algorithm}
\usepackage{algorithmic}
\AtBeginDocument{%
  \providecommand\BibTeX{{%
    \normalfont B\kern-0.5em{\scshape i\kern-0.25em b}\kern-0.8em\TeX}}}

\setcopyright{acmcopyright}
\copyrightyear{2021}
\acmYear{2021}
\acmDOI{10.1145/1122445.1122456}

\acmConference[Chengdu '21]{Chengdu '21: ACM Symposium on Neural
  Gaze Detection}{October  20--24, 2021}{Chengdu, Sichuan}
\acmBooktitle{Chengdu '21: ACM Symposium on Neural Gaze Detection,
  October  20--24, 2021, Chengdu, Sichuan}
\acmPrice{15.00}
\acmISBN{978-1-4503-XXXX-X/18/06}

\begin{document}

\title{RAMS-Trans: Recurrent Attention Multi-scale Transformer for Fine-grained Image Recognition}

\author{Yunqing Hu$^{1,2}$,\quad Xuan Jin$^2$,\quad Yin Zhang$^{1*}$,\quad Haiwen Hong$^{1,2}$, \quad Jingfeng Zhang$^{1,2}$, \quad Yuan He$^{2}$, \quad Hui Xue$^{2}$}

\makeatletter
\def\authornotetext#1{
\if@ACM@anonymous\else
    \g@addto@macro\@authornotes{
    \stepcounter{footnote}\footnotetext{#1}}
\fi}
\makeatother
\authornotetext{Corresponding author.}

\affiliation{
 \institution{\textsuperscript{\rm 1}College of Computer Science and Technology, Zhejiang University \city{Hangzhou} \country{China}}
 \institution{\textsuperscript{\rm 2}Alibaba Group \city{Hangzhou} \country{China}}
 }
\email{{yunqinghu, zhangyin98, honghaiwen96, zhjf}@zju.edu.cn}
\email{{jinxuan.jx, heyuan.hy, hui.xueh}@alibaba-inc.com}

\def\authors{Yunqing Hu, Xuan Jin, Yin Zhang, Haiwen Hong, Jingfeng Zhang, Yuan He, Hui Xue}

\renewcommand{\shortauthors}{Yunqing Hu et al.}

\begin{abstract}
  In fine-grained image recognition (FGIR), the localization and amplification of region attention is an important factor, which has been explored a lot by convolutional neural networks (CNNs) based approaches. The recently developed vision transformer (ViT) has achieved promising results on computer vision tasks. Compared with CNNs, Image sequentialization is a brand new manner. However, ViT is limited in its receptive field size and thus lacks local attention like CNNs due to the fixed size of its patches, and is unable to generate multi-scale features to learn discriminative region attention. To facilitate the learning of discriminative region attention without box/part annotations, we use the strength of the attention weights to measure the importance of the patch tokens corresponding to the raw images. We propose the recurrent attention multi-scale transformer (RAMS-Trans), which uses the transformer's self-attention to recursively learn discriminative region attention in a multi-scale manner. Specifically, at the core of our approach lies the dynamic patch proposal module (DPPM) guided region amplification to complete the integration of multi-scale image patches. The DPPM starts with the full-size image patches and iteratively scales up the region attention to generate new patches from global to local by the intensity of the attention weights generated at each scale as an indicator. Our approach requires only the attention weights that come with ViT itself and can be easily trained end-to-end. Extensive experiments demonstrate that RAMS-Trans performs better than concurrent works, in addition to efficient CNN models, achieving state-of-the-art results on three benchmark datasets.
\end{abstract}




\maketitle

\section{Introduction}
Fine-grained image recognition (FGIR) has been a challenging problem. Most of the current methods are dominated by convolutional neural networks (CNNs). Unlike conventional image classification problems, FGIR has the problem of large intra-class variance and small inter-class variance. Therefore, FGIR methods need to be able to identify and localize region attention in an image that is critical for classification. There is a class of methods called part-based methods \cite{Mask-CNN, Pose} for FGIR, and some of them use additional supervised information such as bounding box/part annotations to locate key regions. However, labeling bounding boxes/part annotations is a labor-intensive task that requires a lot of resources. How to be able to use the effective information generated by the model itself for region attention localization and amplification is one of the research directions that FGIR has to face. 

The effectiveness of CNNs needs no further explanation here. However, we need to emphasize again that one of the key aspects of CNNs that make them effective is their translation invariance and local feature representation capability. CNNs are continuously downsampled as their depth increases, yet the receptive field of the model increases, so that both global and local information of the feature map can be utilized. For example, in networks such as VggNet \cite{vgg} and ResNet \cite{ResNet}, the receptive field of underlying convolutional is smaller and has more local information, while the receptive field of higher convolutional is larger and has more global information. The works \cite{Multi-attention} and \cite{learn2pay} use this characteristic for FGIR. Some works exploit the attention properties of the feature maps of CNN itself, such as \cite{MMAL-Net} and \cite{SCDA} that exploit the attention maps of image features to select region attentions.
Transformer \cite{attention_is} has gradually transformed from a research hotspot in NLP \cite{Bert, XL, XLNet} to  CV tasks \cite{ViT, deform} in recent years. The proposal of vision transformer (ViT) has brought a new shock to computer vision and aroused the research interest in image sequentialization within the community. ViT flattens the segmented image patches and transforms them into patch tokens. Similar to character sequences in NLP, those tokens will be sent to the multi-head self-attention mechanism for training. Since patch tokens are position-agnostic, position embedding will be added to serve the purpose of adding spatial information. However, when ViT encounters FGIR, there are two main problems need to be solved. First, The model processes all the patch tokens at once, and when the complexity of the dataset increases, such as when the image resolution is high or when an image has a cluttered background, the model may not be able to effectively capture the region attention carried in the patch tokens. Conversely, when the image resolution is low, this fixed patch size is more likely to make the model lose local information. Second, ViT differs from CNNs in that the length of patch tokens does not change as its encoder blocks increases, thus the receptive field of the model cannot be effectively extended.
Therefore, for FGIR, we can do much more than just feed the flattened raw image patches into the Transformer. If we refer to the characteristics of CNNs and introduce attention to local regions in the model, that is, extend the effective receptive field, the recognition performance of the model is likely to be further improved. So then we encounter a very important question, how to explore and discover local information in ViT which focuses on using global information?  The latest TransFG \cite{TransFG} gives us a pretty good answer, which is to take advantage of ViT's inherent attention weights. Both in the NLP domain of transformer and ViT training, most of the works simply require the use of the last layer of classification token information, discarding the seemingly accessory attention weights. TransFG multiplies all the attention weights before the last transformer layer to get the importance ranking of patch tokens, and then concatenate the selected tokens along with the global classification token as input sequence to the last transformer layer. However, this kind of hard attention filtering is easy to fail in two cases, one is in the case of small image resolution, and the other is in the case of the high complexity of the dataset. In the former case, a lot of important local information is not easily available, and if most of the tokens information has to be filtered out at this time, it is likely to lose classification performance. In the latter case, a model can easily make wrong judgments based on improper token information when the attention mechanism fails.

Through preliminary visualization experiments, We find that the strength of the attention weights can be intuitively correlated with the extent to which the patches contain the target object. To this end, we propose the recurrent attention multi-scale transformer (RAMS-Trans), which uses the transformer's self-attention mechanism to recursively learn discriminative region attention in a multi-scale manner. Specifically, at the core of our approach lies the proposed dynamic patch proposal module (DPPM) which aims to adaptively select the most discriminative region for each image. The DPPM starts with the complete image patches and scales up the region attention to generate new patches from global to local by the intensity of the attention weights generated at each scale as an indicator. The finer scale network takes as input the tendency regions scaled up from the previous scale in a cyclic manner. The following contributions are made in this paper:
\begin{itemize}
\item We reformulate the FGIR problem from a sequence to sequence learning perspective and design a new visual Transformer architecture namely recurrent attention multi-scale transformer (RAMS-Trans). It combines the advantages of CNNs in expanding receptive field, strengthening locality, and the advantages of Transformers in utilizing global information.
\item As an instantiation, we exploit the transformer framework, specifically, to the use of multi-head self-attention weights to locate and zoom in on regions of interest, to implement our fully attentive feature representation by sequentializing images.
\item Extensive experiments show that our RAMS-Trans model can learn superior feature representations as compared to traditional CNNs and concurrent work on three popular FGIR benchmarks (CUB-200-2011, Stanford Dogs, and iNaturalist2017).
\end{itemize}

\section{Related Work}
\subsection{CNN based Fine-grained image recognition}
FGIR can be divided into the following three directions, localization-classification sub-networks, end-to-end feature encoding, and external information, of which the first two directions are the main content of this section. The first method is classified as strongly \cite{Pose, Mask-CNN, HSnet} or weakly supervised \cite{StackedLSTM} according to whether it utilizes bounding box/part annotations information. This class of methods locates key component regions by training supervised or weakly supervised localization sub-networks. Then, the classification sub-network uses the fine-grained region information captured by the localization sub-network to further improve the classification capability. Mask-CNN \cite{Mask-CNN} is based on part annotations and uses FCN to localize key parts (head, torso) to generate a mask with weighted object/part. However, the acquisition of part annotation can add additional and high markup costs. Many methods use attention mechanisms to set specific sub-network structures so that classification can be done using only image-level annotations. The second type of approach usually designs end-to-end models that encode discriminative features as higher-order information. From Bilinear-Pooling \cite{Bilinear-CNN} to compact kernel pooling \cite{Compact}, many works use different methods such as designing kernel modules \cite{Kernel_Pooling} or special loss functions \cite{DBTNet} to reduce the dimensionality of higher-order features. However, these methods have difficulty in obtaining fine variance from the global feature view and hardly surpass the previous method.

An approach very close to our work is RA-CNN \cite{RA-CNN}, and the common denominator is the learning of regional features under the action of two scales. However, we have the following two key differences from RA-CNN. First, we don't need additional parameters to learn the coordinates of the regions, and we only need to rely on the attention weights attached to the transformer training for the region attention learning. Second, we do not need to force the accuracy of scale 2 to be higher than scale 1, and we are letting the two scales learn from each other and jointly improve the accuracy.
\subsection{Transformer in Vision} 
Inspired by the Transformer \cite{attention_is} for NLP tasks \cite{Bert, XLNet, XL}, a large number of models have recently emerged that rely heavily on the Transformer for computer vision \cite{ViT, deform}. \cite{end2end} and \cite{deform} are the earlier works to apply transformer to object detection. ViT \cite{ViT} is the first work to transform a 2D image into an 1D patch tokens, which will be fed into the subsequent transformer layers for training, achieving an accuracy rate comparable to that of CNNs for image recognition. DeiT \cite{Deit} enhances ViT \cite{ViT} by introducing a simulation token and employs knowledge distillation to simulate the output of CNN teachers to obtain satisfactory results without training on large-scale datasets. SETR \cite{SETR} proposes a pure self-attention-based encoder to perform semantic segmentation.

The most related work is TransFG \cite{TransFG} which also leverages attention for FGIR. However, there is a key difference. We take the use of attention weights for amplification and reuse of region attentions, while TransFG only filters the patch tokens in the last layer of the Transformer. Second, we propose a recurrent structure to extract and learn multi-scale features for better visual representation. Our model is superior in various image resolutions and large-scale FGIR datasets (see Section 4).

\section{Approach}
Our approach (RAMS-Trans) is built on top of the vision transformer (ViT \cite{ViT}), so we first present a brief overview of ViT and then describe our approach for learning multi-scale features for FGIR.

\subsection{Preliminaries: Vision Transformer}
\textbf{Image Tokenization.} The innovation and key to ViT are that it processes a 2D image into a string-like 1D sequence and then feeds it into blocks stacked by the standard Transformer's encoder. Specifically, ViT reshapes the image $x$, $x \in \mathbb{R}^{H \times W \times 3}$, with certain patch size, into a 2D sequence of patches $x_p$,  $x_p \in \mathbb{R}^{N \times (P \times P \times 3)}$, where H, W are the height and width of the raw image respectively, 3 is the number of channels of the raw RGB image, and P is the artificially set patch size used to split the image. In ViT, the size of $P$ is usually 16 or 32. $N$ is the total number of patches split into, $N = H \times W /P^2$. Then ViT maps the vectorized patches $x_p$ into a latent $C$-dimensional embedding space using a trainable linear projection, to obtain the patch tokens $x_{patch}$, where $x_{patch} \in \mathbb{R}^{N\times C}$. Similar to BERT, ViT also initializes the class tokens (CLS) for final classification in the tokenization phase, which will be concatenated with the patch tokens and then sent to the subsequent transformer layers. In addition, since the patch tokens input to the subsequent transformer are position-agnostic, and the image processing depends on the spatial information of each pixel, ViT adds the position embedding to each patch, which can be continuously learned in the subsequent training process:
\begin{equation}
    x0 = [x_{cls}||x_{patch}] + x_{pos}
\end{equation}
where $x_{cls} \in \mathbb{R}^{1\times C}$ and $x_{pos} \in \mathbb{R}^{(1+N) \times C}$ are the CLS and the position embedding respectively.
However, image tokenization in FGIR with a fixed patch size may have two problems: (1) The model processes all the patch tokens at once, and when the complexity of the dataset increases, e.g., with a cluttered background, the model may not be able to effectively capture the region attention carried in the patch tokens. (2) This kind of fixed patch size makes it easier for the model to lose local information when the image resolution is low.

\textbf{Encoder Blocks.} The main structure of ViT are Blocks, consists of a stack of $L$ Transformer's standard encoder. Each block consists of a multi-head self-attention (MSA) and a feed-forward network (FFN), which consists of two fully connected layers. The output of the $k_{th}$ layer can be expressed as:
\begin{equation}
    y_k = x_k - 1 + MSA(LN(x_k - 1))  
\end{equation}
\begin{equation}
x_k = y_k + FFN(LN(y_k))    
\end{equation}
where $LN(\cdot)$ indicates the Layer Normalization operation \cite{LN}. The uniqueness of CNN's image processing lies in the fact that as the depth of the model increases, the raw images are continuously downsampled, while the receptive field of the model keeps getting larger so that both global and local information of the images can be utilized. What makes ViT different from CNNs is with the increasing number of encoder blocks, the length of patch tokens does not change, and the receptive field of the model cannot be effectively extended, which may affect the accuracy of the model on the FGIR.

\begin{figure}[h]
\centering
\includegraphics[width=0.42\textwidth]{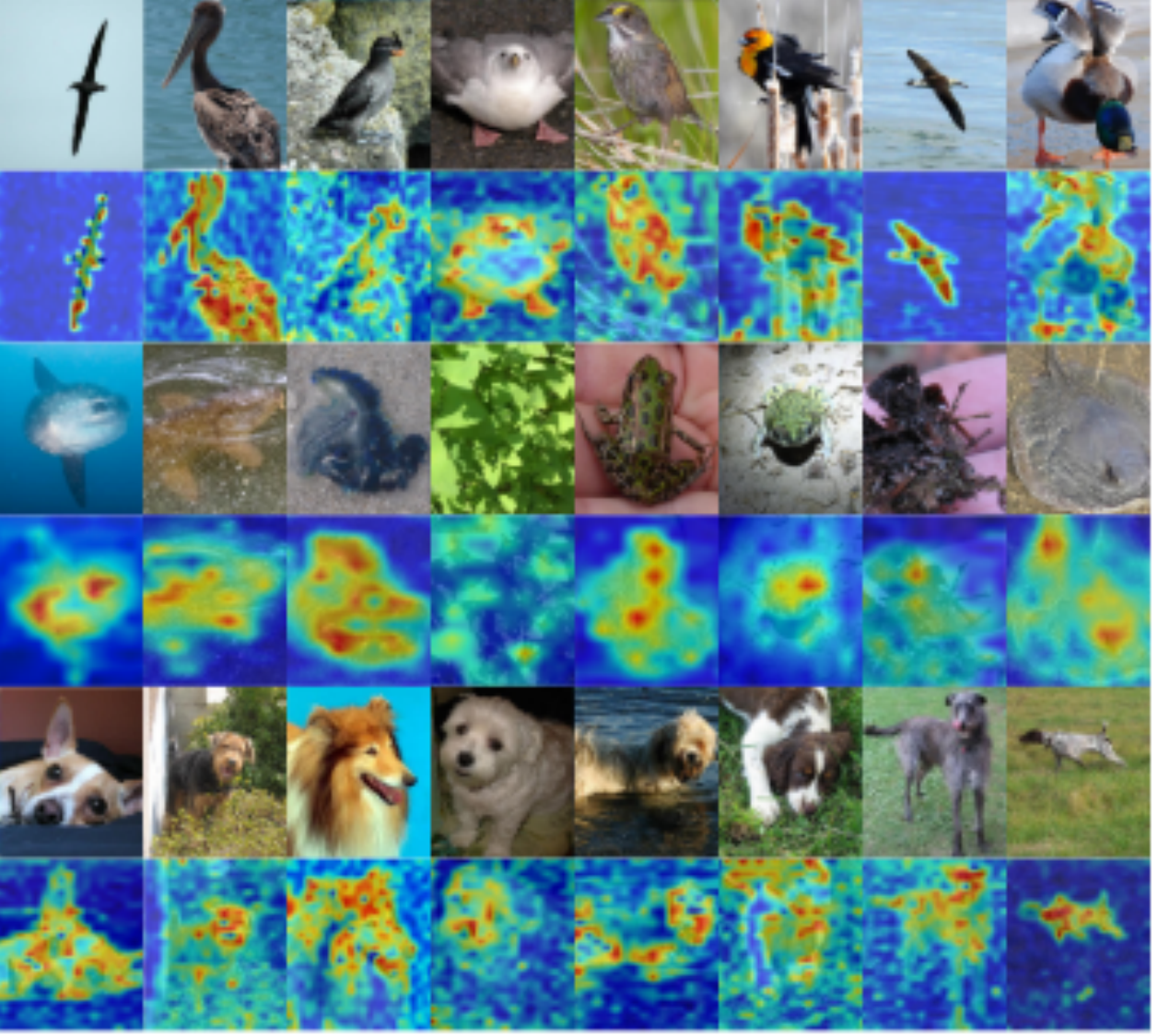}
\caption{Visualization results of Attention weights on CUB-200-2011, iNaturalist2017 and Stanford dogs datasets. The first, the third and the fifth rows are original images, while the second, the fourth and the sixth raws show the raw attention maps. Best viewed in color.}
\end{figure}

\begin{figure*}[t]
\centering
\includegraphics[width=0.8\textwidth]{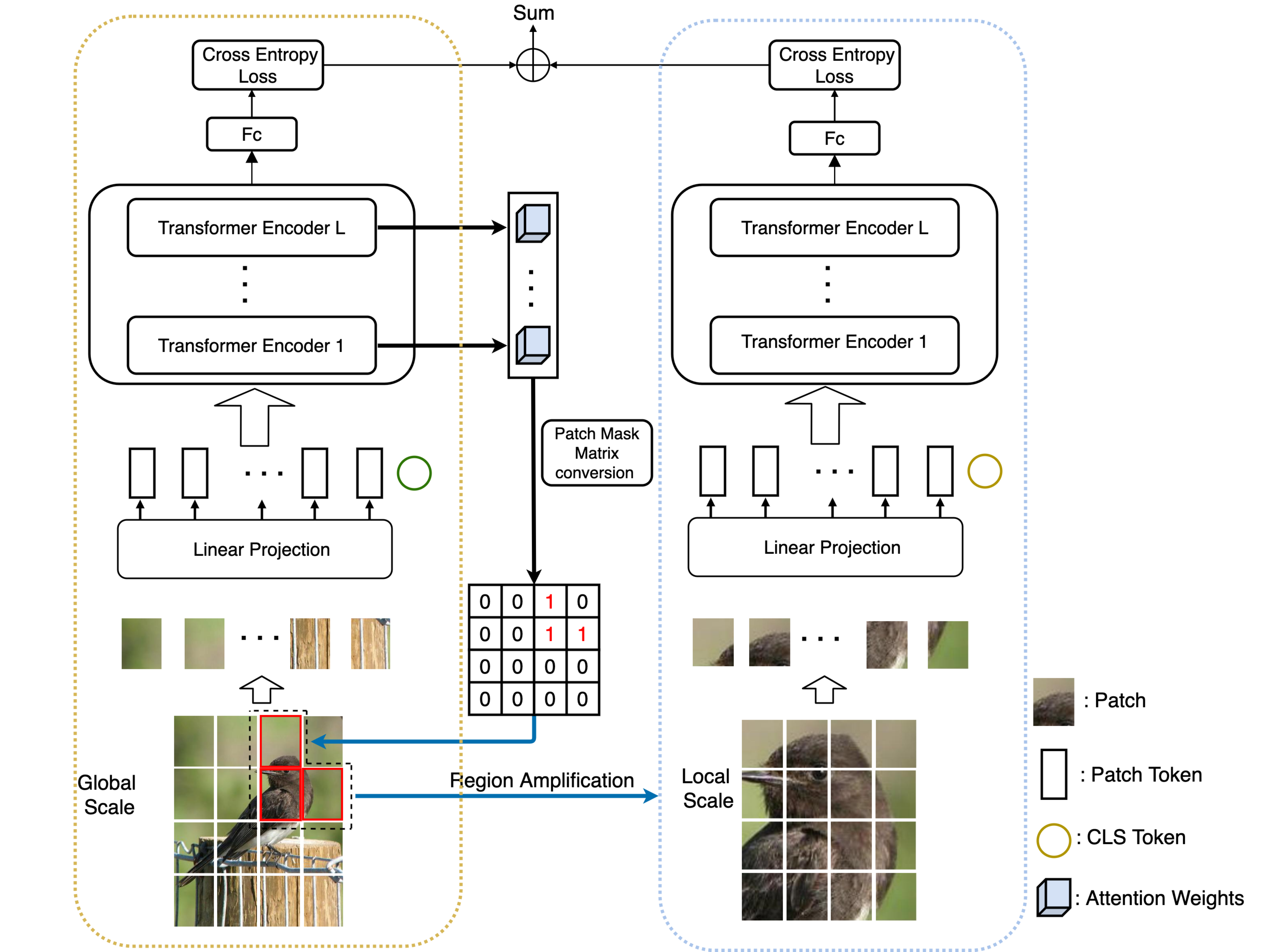}
\caption{The framework of recurrent attention multi-scale transformer (RAMS-Trans). The inputs are from global full-size images to local region attention (from left to right). The attention weights of all transformer layers are aggregated to generate the patch mask matrix, of which red 1 indicates an activated patch. The red box indicates the selected patch. Note that the linear projection, the transformer layers, and Fc (Fully Connection) layers are parameter-sharing, while CLS tokens do not.}
\end{figure*}

\subsection{Proposed Network}
Inspired by SCDA \cite{SCDA} and RA-CNN \cite{RA-CNN}, we propose the recurrent attention multi-scale transformer (RAMS-Trans) to solve the above problems. We take two scales in Figure 2 as an example. First, the model accepts the raw input image $x_1$ and then outputs cross-entropy loss1 and the multi-head self-attention weights $w_l$ of each transformer layer after the computation. Then $w_l$ is fed into DPPM, which firstly outputs the corresponding patches mask matrix on the raw image $x_1$, that is, the binary patch mask matrix, and then gets the coordinate value of the region attention on $x_1$ by the maximum connected region search algorithm according to the matrix. In the second stage, we get the local image input $x_2$ by bilinear interpolation algorithm to zoom in from $x_1$, which will be recomputed through L-layer encoder blocks to get cross-entropy loss2 and multi-head self-attention weights $w_2$.

It is important to emphasize that the core of our approach is how to use the characteristics of ViT itself to find the region attention in the raw image, to break the receptive field limitation of ViT due to the fixed size of patches, and then use the multi-scale image representation to achieve more effective recognition of objects. In CNNs, SCDA uses the fusion of multiple channel feature maps to achieve the localization of objects, from which we are inspired that since ViT processes the raw images into patch tokens for subsequent MSA and FFN calculations, can we thus obtain the importance of each patch corresponding to the raw image $x_1$? Since one of the cores of the Transformer is MSA, then it is natural to think of using self-attention weights to try to accomplish this. We first perform a visual evaluation:

\textbf{Visual Evaluation.} Relying only on the attention weights incidental to the pre-training of ViT, it is possible to accurately locate the object region in the raw images and discard the irrelevant and noisy regions. In Figure 1., we show some images from three datasets CUB-200-2011, Stanford Dogs, and iNaturalist2017. We extract their attention weights using a ViT model pre-trained on ImageNet21k without fine-tuning the target dataset at all, and then visualize them employing CAM \cite{CAM}. In Figure 1 we can see that using only the raw attention weights we can well localize the objects in the raw images and mask the background and noise regions. The above visualization process illustrates that the self-attention weights generated by ViT in the calculation of its MSA mechanism can be correlated to some extent with the positions of the target objects in the raw images.

\textbf{Dynamic Patch Proposal Module.}
Our goal is to adaptively select a varying number of patches from $x_1$ with $N^{1/2}\times N^{1/2}$ patches to recompose $x_2$. We first take out the attention weights of each transformer layer as:
\begin{equation}
W = softmax(\frac{QK^T}{C^{1/2}})= [w_0, w_1, ...,w_L]
\end{equation}
\begin{equation}
w_l = [w_l^1, w_l^2, ...,w_l^K] \quad l\in1,2,...,L
\end{equation}
\begin{equation}
    w_l^i = [w_l^1, w_l^2, ...,w_l^C] \quad i\in1,2,...,K
\end{equation}
where Q, K are Query and Key vectors respectively.
Then we regularize the $w_l$
\begin{equation}
    G_l = \frac{\frac{1}{K}\sum_{k={1}}^{K}w_l^k+E}{\chi}
\end{equation}
where $\chi$ is the regularization factor and $E$ is the diagonal matrix.
\begin{equation}
    \chi = \frac{1}{C}\sum_{c=1}^{C}(\frac{1}{K}\sum_{k={1}}^{K}w_l^k+E)
\end{equation}
Then we propose to integrate attention weights of all previous layers and recursively apply a matrix multiplication to the modified attention weights in all the layers as:
\begin{equation}
    g = \prod_{l=1}^{L}{G_l}
\end{equation}

We calculate the mean value $\overline{g}$ of all the positions in $g$ as the threshold to determine the localization position of the object. In particular, to improve the localization ability and further determine the region attention, we design a magnification factor $\alpha$ as a hyperparameter to increase the threshold:
\begin{align}
    \tilde{M}_{(x,y)} = \begin{cases}
1  \quad if \quad g_{(x,y)} > \alpha \overline{g}\\
0 \quad otherwise \\
\end{cases}
\end{align}
where $\tilde{M}_(x,y)$ is the patch mask matrix and (x, y) is a particular position in these $N^{1/2}\times N^{1/2}$ positions.
Finally, we employ Algorithm 1 to extract the largest connected component of $\tilde{M}$ to localize and zoom in the region attention in the raw image $x_1$.

\subsection{Implementation}
We present our implementation details on loss function as well as scale-wise class token.

    \textbf{Loss Function} In the training phase, our loss function is represented as a multi-task loss $Loss_{total}$ consisting of classification $Loss_{s1}$ and guided $Loss_{s2}$:
\begin{equation}
    Loss_{total} = Loss_{s1} + \lambda Loss_{s2}
\end{equation}
which are complementary to each other. $\lambda$ is the coefficient to balance the weight between the two losses, which we take as $1.0$ in the experiments. $Loss_{s1}$ represents the fine-grained classification loss of scale1 and $Loss_{s2}$ is the guided loss which is designed to guide the mode to select the more discriminative regions. These two losses work together in the backpropagation process to optimize the performance of the model. It enables the final convergent model to make classification predictions based on the overall structural characteristics of the object or the characteristics of region attention. During the testing phase, we removed the scale 2 to reduce a large number of calculations, so our approach will not take too long to predict in practical applications. 

\textbf{Scale-wise Class Token} In Sec 3.1 we have described how the class token is generated and its role, which is mainly to exchange information with the patch tokens and finally to feed the class information to the classification layer. However, in our framework, the region attention of the raw image will be positioned and enlarged, and thus the patch tokens will be different between scales, which may affect the final classification performance if the class token is shared between scales. We, therefore, propose the scale-wise class token, i.e., different class-tokens are used to adapt patch tokens of different scales:
\begin{equation}
    x0 = [x_{cls1}||x_{cls2}||x_{patch}] + x_{pos}
\end{equation}
We demonstrate the effectiveness of this design in subsequent experiments with different resolutions.

\begin{algorithm}[tb]
\caption{Finding Connected Components in Binary Patch Mask Marix}
\label{alg:algorithm}
\textbf{Require}: A binary matrix: $M$;
\begin{algorithmic}[1] 
\STATE{Select a patch $m$ as the starting point;}
\WHILE{True} 
    \STATE{Use a flood-fill algorithm to label all the patches in the connected component containing $m$};
    \IF{All the patches are labeld}
    \STATE{Break;}
    \ENDIF
\STATE Search for the next unlabeled patch as $m$;
\ENDWHILE
\RETURN{Connectivity of the connected components, and their corresponding size (patches numbers)}
\end{algorithmic}
\end{algorithm}

\section{Experiments}
In this section, we describe our experiments and discuss the results. We first present three datasets and then present our specific experimental details and results for each dataset respectively. Finally, we conduct detailed ablation experiments on our approach to investigate the impact of the components on FGIR in more depth. Note that all our results are reported as accuracies and are compared with the latest methods.

\textbf{Datasets.} A total of three benchmark datasets are presented in our experiments, namely CUB-200-2011 \cite{CUB}, Stanford Dogs \cite{Stanford_dogs} and iNaturalist2017 \cite{iNat}. CUB-200-2011 is a fine-grained dataset on bird classification. In addition to labels, it also contains bounding box/part annotations which are useful for classification. Stanford Dogs contains images of 120 breeds of dogs from around the world. iNaturalist2017 is a large-scale FGIR dataset containing over 5,000 species of animals and plants.

\subsection{Results on CUB-200-2011}

\textbf{Implementation Details.} We load the model weights from the official ViT-B\_16 model pre-trained on ImageNet21k. In all experiments, we used the SGD optimizer to optimize with an initial learning rate of 0.03 and a momentum of 0.9. We use weight decay 0.  We use cosine annealing to adjust the learning rate with batch size 16. The model is trained for a total of 10,000 steps, of which the first 500 steps are warm-up. We resize the input images by scaling the shortest side to 600 and randomly crop a region of $448 \times 448$ for training. In the test, we use center crop to change the image size to $448 \times 448$. We split the image into patches as in the ViT, with the patch size is $16 \times16$. The hyperparameter $\alpha$ is chosen to be 1.3. We complete the construction of the whole model using Pytorch and run all experiments on the Tesla V-100 GPUs.

\textbf{Comparison with state-of-the-art methods.} The classification accuracies of CUB-200-2011 are summarized in Table 1. All previous FGIR methods test their performance on this dataset. As can be seen in Table 1, our approach outperforms all CNN-based methods and TransFG's PSM module and achieves state-of-the-art performance. Although ViT itself achieves good performance on this dataset, with the addition of our DPPM module, it achieves a further 0.7\% improvement, which is rare for the Vit-B\-16 model which has been adequately pre-trained on a large-scale dataset.

\textbf{Visualization.}
    In order to visually analyze the selected regions from raw images for our DPPM, we present the amplified regions in the left part of Figure 3. The first, third, and fifth rows are the raw images, and the second, fourth and sixth rows are the visualization of the local images after the proposed patches have been amplified. From Figure 3 we can clearly see that DPPM has amplified the most discriminative regions of birds in each category, such as the head and the bill.

\subsection{Results on iNaturalist2017}

\textbf{Implementation Details.} To fully validate the effectiveness of our RAMS-Trans, we load the officially provided ViT-B\-16 model pre-trained on ImageNet21k. In all experiments, we use the SGD optimizer to optimize with an initial learning rate of 0.005 and a momentum of 0.9. We use weight decay 0. We use cosine annealing to adjust the learning rate with batch size 16. The model is trained for a total of 1e6 steps, of which the first 500 steps are warm-up. To align with TransFG, we resize the input images by scaling the shortest side to 448 and randomly crop a region of $304 \times 304$ for training. In the test, we use center crop to change the image size to $304 \times 304$. We still split the image into $16 \times 16$ patches with non-overlap and use a hyperparameter alpha of 1.2.

\textbf{Comparison with state-of-the-art methods.} Table 2 summarizes our results and compares them with the CNN-based state-of-the-art methods and TransFG. Our approach outperforms ResNet152 by 9.5\% and outperforms all CNN-based methods. With the pre-trained model loaded, our method can achieve an improvement of 1.5\% higher than the baseline. It is worth note that for a fair comparison, we report in Table 2 both the result obtained when we run the PSM module of TransFG in our code environment under the non-overlap setting. It can be seen that the DPPM module of our approach outperforms the PSM module of TransFG by 2.4\% with the same loading of the ViT-B\_16 pre-trained model.

\textbf{Visualization.}
In order to visually analyze the selected regions from raw images for our DPPM, we present the amplified regions in the right part of Figure 3. The first, third, and fifth rows are the raw images, and the second, fourth, and sixth rows are the visualization of the local images after the proposed patches have been amplified. From Figure 3 we can clearly see that our RAMS-Trans successfully captures the most discriminative regions for an object, i.e., head, eyes for Amphibia; fins for Actinopterygii; thallus for Protozoa. 

\subsection{Results on Stanford Dogs}

\textbf{Implementation Details.} We load the model weights from the official ViT-B\_16 model pre-trained on ImageNet21k. In all experiments, we use the SGD optimizer to optimize with an initial learning rate of 0.003 and a momentum of 0.9, aligning with TransFG. We use weight decay 0. We use cosine annealing to adjust the learning rate with batch size 16. The model is trained for a total of 20,000 steps, of which the first 500 steps are warm-up. We resize the input images by scaling the shortest side to 448 and randomly crop a region of $224 \times 224$ for training. In the test, we use center crop to change the image size to $224 \times 224$. We split the image into patches as in the ViT, with the patch size is $16 \times 16$. The hyperparameter $\alpha$ is chosen to be 1.0.

\begin{table}[h]
\caption{Comparison of our RAMS-Trans with existing state of the arts methods on CUB-200-2011.}
\begin{center}
\begin{tabular}{c|rr}
\hline
Method
&\multicolumn{1}{c}{Backbone}
&\multicolumn{1}{c}{Acc.(\%)}
\\
\hline
ResNet-50 \cite{ResNet} & ResNet-50 & 84.5\\
RA-CNN \cite{RA-CNN}& VGG-19 & 85.3\\
GP-256 \cite{GP-256}& VGG-16 & 85.8 \\
MaxExt \cite{MaxEnt}& DenseNet-161 & 86.6\\
DFL-CNN \cite{DFL-CNN}& ResNet-50 & 87.4\\
NTS-Net \cite{NTS-Net}& ResNet-50 & 87.5\\
Cross-X \cite{Cross-X}& ResNet-50 & 87.7\\
DCL \cite{DCL}& ResNet-50 & 87.8\\
CIN \cite{CIN}& ResNet-101 &88.1\\
DBTNet \cite{DBTNet}& ResNet-101 &88.1\\
ACNet \cite{ACNet}& ResNet-50 &88.1 \\
S3N \cite{S3N}& ResNet-50 & 88.5\\
FDL \cite{FDL}& DenseNet-161 & 89.1\\
PMG \cite{PMG}& ResNet-50 & 89.6\\
API-Net \cite{API-Net}& DenseNet-161 & 90.0\\
StackedLSTM \cite{StackedLSTM}& GoogleNet & 90.4\\
MMAL-Net \cite{MMAL-Net}& ResNet-50 & 89.6\\
\hline
ViT  \cite{ViT}& ViT-B\_16 & 90.6\\
TransFG \& PSM \cite{TransFG}& ViT-B\_16 & 90.9\\
RAMS-Trans & ViT-B\_16 & \textbf{91.3}\\
\hline
\end{tabular}
\end{center}
\label{tab_base_model}
\end{table}

\begin{table}[t]
\caption{Comparison of our RAMS-Trans with existing state of the arts methods on iNaturalist2017.}
\centering
\begin{tabular}{c|c|c}
\hline
Method
&Backbone
&iNaturalist2017
\\
\hline
Resnet152 \cite{ResNet}
&ResNet152
&59.0
\\
SSN \cite{SSN}
&ResNet101
&65.2

\\
Huang et al. \cite{Interpretable}
&ResNet101
&66.8

\\
IncResNetV2 \cite{Inception-v4}
&InResNetV2
&67.3

\\
TASN \cite{TASN}
&ResNet101
&68.2

\\
\hline

ViT \cite{ViT}
&ViT-B\_16
&67.0

\\
TransFG \&PSM \cite{TransFG}
&ViT-B\_16
&66.6
\\
RAMS-Trans
&ViT-B\_16
&\textbf{68.5}

\\
\hline
\end{tabular}
\label{tab_acc_size}
\end{table}

\begin{table}[t]
\caption{Comparison of our RAMS-Trans with existing state of the arts methods on Stanford Dogs.}
\centering
\begin{tabular}{c|c|c}
\hline
Method
&Backbone
&Stanford Dogs
\\
\hline
MaxEnt \cite{MaxEnt}
&DenseNet-161
&84.9
\\
FDL \cite{FDL}
&DenseNet-161
&84.9

\\
RA-CNN \cite{RA-CNN}
&VGG-19
&873

\\
SEF \cite{SEF}
&ResNet-50
&88.8

\\
Cross-X \cite{Cross-X}
&ResNet-50
&88.9
\\
API-Net \cite{API-Net}
&ResNet-101
&90.3

\\
\hline

ViT \cite{ViT}
&ViT-B\_16
&92.2

\\
TransFG \& PSM \cite{TransFG}
&ViT-B\_16
&90.0

\\
RAMS-Trans
&ViT-B\_16
&\textbf{92.4}

\\
\hline
\end{tabular}
\label{tab_acc_size}
\end{table}

\begin{table}
\caption{Ablation experiments on DPPM with different input resolutions.}
\begin{center}
\begin{tabular}{c|rrrrrr}
\hline
Resolution & 192 & 224 & 256
&288 & 320 \\
\hline\hline
ViT & 81.5 & 85.3 & 87.4 & 88.3 & 88.9 \\
TransFG \& PSM \cite{TransFG} & 81.6 & 84.9 & 86.8 & 88.0 & 89.1 \\
\hline
RAMS-Trans & \textbf{82.2} & \textbf{86.3} & \textbf{88.1} & \textbf{88.9} & \textbf{89.7} \\
\hline
\end{tabular}
\end{center}
\label{tab_ablation}
\end{table}

\textbf{Comparison with state-of-the-art methods.} The classification results of Stanford Dogs are summarized in Table 3. As can be seen, our approach outperforms all CNN-based methods and TransFG and achieves state-of-the-art results. It is important to note that for TransFG, the DPPM module of ours is aligned with its PSM module. Although they are both hard attention mechanisms, our approach is softer than its simple token filtering way because our approach extends the effective receptive field of the raw images, thus the classification performance is also better.

\textbf{Visualization.}
 In order to visually analyze the selected regions from raw images for our DPPM, we present the amplified regions in the center part of Figure 3. The first, third and fifth rows are the raw images, and the second, fourth and sixth rows are the visualization of the local images after the proposed patches have been amplified. Figure 3 conveys that the proposed regions do contain more fine-grained information of dogs in each category such as the ears, eyes, and fur.
 
 \begin{figure*}[h]
\centering
\includegraphics[width=1.0\textwidth]{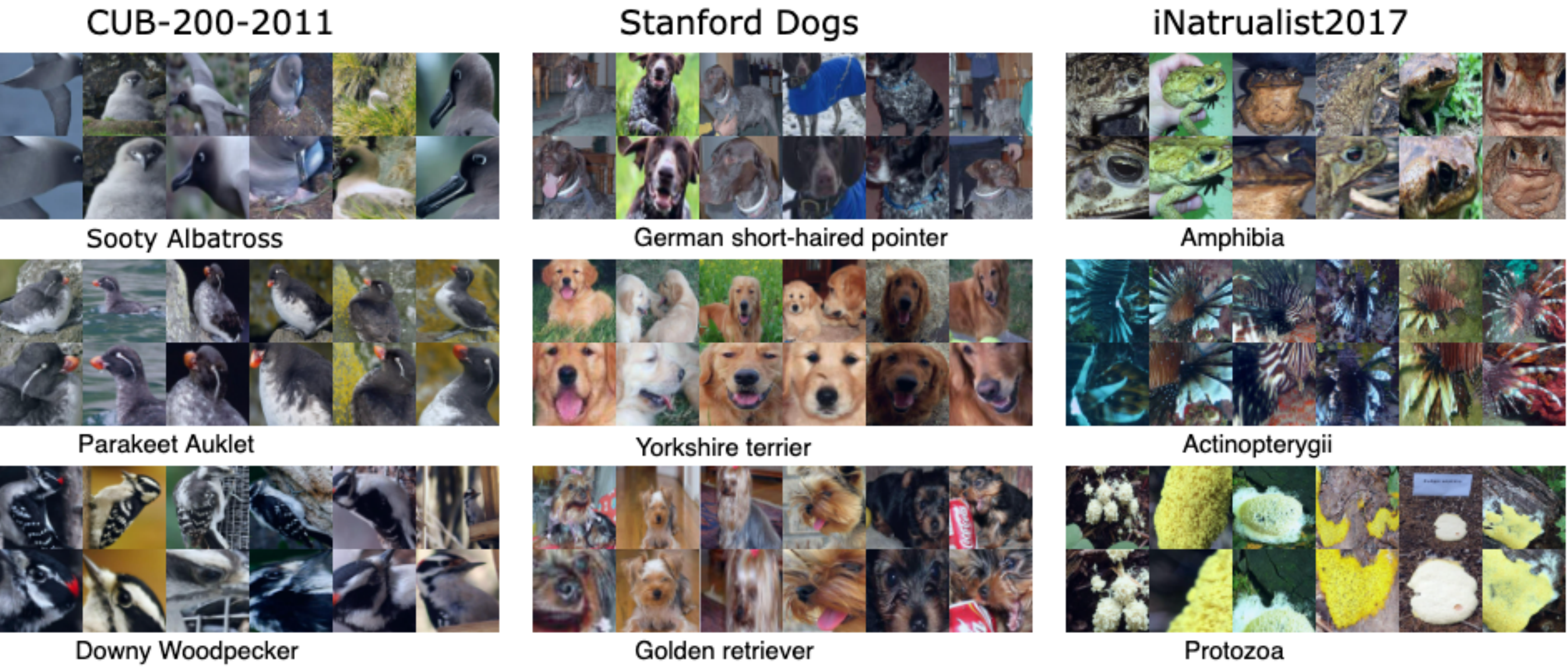}
\caption{Attention localization at the local scale for CUB-200-2011, Stanford Dogs and iNaturalist2017. The regions (in second row of each category) learned from multiple image samples, represent consistent region attention for a specific fine-grained category, which are discriminative to classify this category from others.}
\end{figure*}

\begin{table}[h]
\caption{Impact of Threshold $\alpha$ on CUB-200-2011 dataset.}
\begin{center}
\begin{tabular}{c|rr}
\hline
Approach
&\multicolumn{1}{c}{Value of $\alpha$}
&\multicolumn{1}{c}{Accuracy(\%)}
\\
\hline
RAMS-Trans & 1.1 & 90.9\\
RAMS-Trans & 1.2 & 91.2\\
RAMS-Trans & 1.3 & \textbf{91.3}\\
RAMS-Trans & 1.4 & 91.0\\
\hline
\end{tabular}
\end{center}
\label{tab_base_model}
\end{table}

\subsection{Ablation Experiments}
We conduct ablation experiments on our RAMS-Trans framework to analyze how its variants affect the FGIR results. All ablation studies are done on CUB-200-2011 dataset while the same phenomenon can be observed on other datasets as well.

\textbf{Hyperparameters.} Since DPPM in our approach requires the choice of hyperparameter $\alpha$, we experimentally investigate the effect of the threshold $\alpha$ on the classification performance. We set 4 sets of alpha values between 1.1 and 1.4 with 0.1 intervals, and the results of all experiments are summarized in Table 5. It can be seen that as the $\alpha$ value increases from 1.1 to 1.4, the recognition accuracy first increases and then decreases. The best performance is obtained when $\alpha$ is 1.3. Based on this, we can get the following analysis: when the $\alpha$ is small, the DPPM will crop as many patches in the raw images as possible, resulting in many non-critical regions being fed into the model again, while when the $\alpha$ is large, the DPPM will crop as few patches in the raw images as possible, losing many critical regions to some extent. Both of these situations will lead to a decrease in classification accuracy, thus it is important to choose a suitable threshold value.

\begin{figure}[h]
\centering
\includegraphics[width=0.38\textwidth]{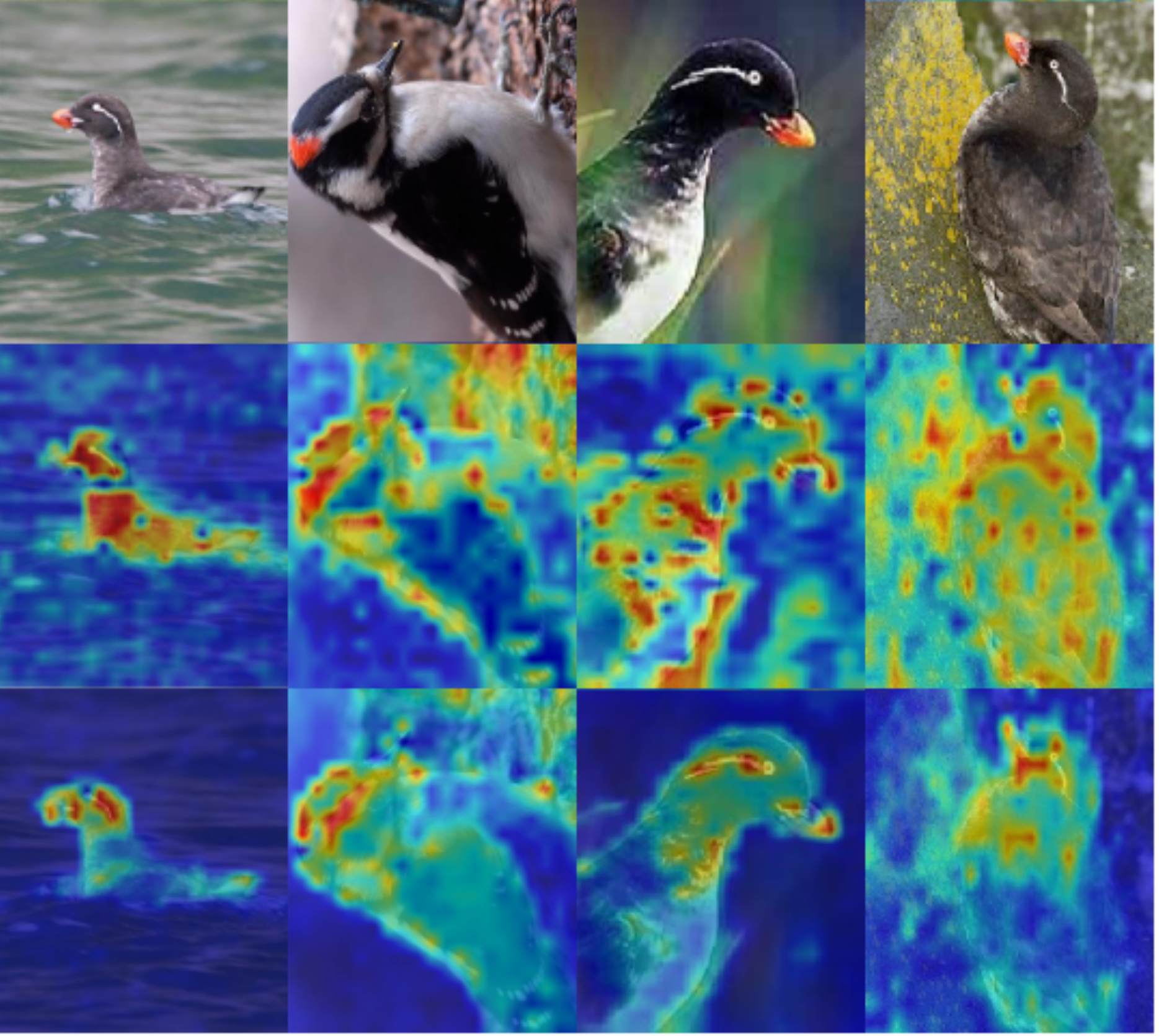}
\caption{An illustration of learning discriminative details by RAMS-Trans. The first row is the original images, the second raw shows the attention maps generated from raw attention weights, and the third raw shows the attention maps generated from trained attention weights.}
\end{figure}

\begin{table}
\caption{Ablation experiments on CLS token with different input resolutions}
\begin{center}
\begin{tabular}{c|rrrrrrr}
\hline
Resolution & 192 & 224 & 256
&288 & 320 &448\\
\hline\hline
scale-sharing & 82.2 & 85.9 & 87.8 & 88.5 & \textbf{89.8} &91.2\\
scale-wise & \textbf{82.2} & \textbf{86.3} & \textbf{88.1} & \textbf{88.9} & 89.7 &\textbf{91.3}\\
\hline
\end{tabular}
\end{center}
\label{tab_ablation}
\end{table}

\begin{table}
\caption{Ablation experiments on patch proposal way.}
\begin{center}
\begin{tabular}{c|cc}
\hline
Method & Patch Proposal & Accuracy(\%) \\
\hline\hline
ViT \cite{ViT} & Bounding Box  & 89.2 \\
RAMS-Trans & Bounding Box  & 90.5 \\
RAMS-Trans & Random & 90.9 \\
RAMS-Trans & DPPM  & \textbf{91.3} \\
\hline
\end{tabular}
\end{center}
\label{tab_ablation}
\end{table}

\begin{table}
\caption{Ablation experiments on patch size.}
\begin{center}
\begin{tabular}{c|cc}
\hline
Method & Patch Size & Accuracy(\%) \\
\hline\hline
ViT \cite{ViT} & $16  \times 16$ & 90.6 \\
TransFG \& PSM \cite{TransFG} & $16  \times 16$ & 90.9 \\
RAMS-Trans & $16  \times 16$ & \textbf{91.3} \\
\hline
ViT & $32  \times 32$  &  88.4\\
TransFG \& PSM \cite{TransFG} & $32  \times 32$  &  88.9\\
RAMS-Trans & $32 \times 32$  & \textbf{89.1} \\
\hline
\end{tabular}
\end{center}
\label{tab_ablation}
\end{table}

\textbf{Compared to bounding box annotations.} In order to further investigate the effectiveness of RAMS-Trans in selecting region attention, we replace our DPPM with the bounding box coordinates that come with the CUB-200-2011 dataset, keeping other settings unchanged, and conduct a comparison experiment. In Table 7, it can be seen that the experiments with a bounding box that utilized the supervised information are instead lower than the baseline. We also compare and analyze these two sets of comparative experiments by zooming in on the local images obtained from the raw images. From Table 7, we can analyze and obtain that using object annotations limits the performance since human annotations only give the coordinates of important parts rather than the accurate discriminative region location. Figure 4 visualizes the change of the feature maps before and after training. It can be seen that the activation region tends to be localized from the original global.

\textbf{Influence of image resolution.} Image resolution is a critical factor for image classification, and there are very many applications in real industrial scenarios. In order to investigate the performance improvement of RAMS-Trans for different image resolutions, we set five different image resolutions for comparison experiments, which are $224 \times 224$, $256 \times 256$, $288 \times 288$ and $320 \times 320$. Meanwhile, we conduct experiments on the PSM of TransFG with the same settings. As shown in Table 4, our method exceeds the baseline at each set of resolutions, while the PSM module loses performance.

\textbf{Compared to random proposal method.} In order to prove the effectiveness of our proposed DPPM module, we add the experiment of random sampling. Instead of the DPPM module, we use randomly generated coordinate points. As shown in Table 7, the DPPM can improve 0.4\% over the random sampling method.

\textbf{Influence of scale-wise class token.} To verify the validity of the scale-wise class token, we add scale-shared class token experiments to experiments 4. As can be seen from Table 6, the experimental results with scale-wise class token are better than those scale-wise class token in most resolution cases.

\textbf{Influence of patch size} To analyze the effect of different patch sizes on FGIR and the effectiveness of our approach at different patch sizes, we load the official ViT-B\_32 model provided for pre-training on ImageNet21k. The experimental results are summarized in Table 8. For a fair comparison, we report in Table 8 both the results obtained when we run the PSM module of TransFG in our code environment under the non-overlap setting. It can be seen that when the patch size is 32, the performance of ViT as a baseline drops by 2.1\% compared to when the patch size is 16. We can analyze that the larger the patch size is, the more sparsely the raw image is split, and the local information is not well utilized, so the performance of FGIR is degraded, yet our approach still exceeds the baseline by 0.7\%. 

\section{Conclusion}
In this paper, we propose a new recurrent attention multi-scale transformer (RAMS-Trans) architecture that combines the advantages of CNNs in expanding the effective receptive field, strengthening locality, and the advantages of Transformers in utilizing global information. Without bounding box/part annotations and additional parameters, RAMS-Trans uses the transformer's self-attention weights to measure the importance of the patch tokens corresponding to the raw images and recursively learns discriminative region attention in a multi-scale manner. Last but not least, our approach can be easily trained end-to-end and achieves state-of-the-art in CUB-200-2011, Stanford Dogs, and the large-scale iNaturalist2017 datasets. The future work is how to locate region attention more precisely to further improve the classification accuracy.

\bibliographystyle{ACM-Reference-Format}
\bibliography{FG.bib}
\end{document}